\documentclass[conference]{IEEEtran}
\pdfoutput=1 
\usepackage{times}
\usepackage{epsfig}
\usepackage{graphicx}
\usepackage{amsmath,amsfonts,amssymb}
\usepackage{balance}
\usepackage[font=footnotesize,labelfont=bf]{caption}
\usepackage[font=footnotesize]{subcaption}
\graphicspath{{./figs/}}

\usepackage{parskip}
\usepackage[pagebackref=true,breaklinks=true,letterpaper=true,colorlinks,bookmarks=false]{hyperref}

\begin{document}
\IEEEoverridecommandlockouts
% Title.
% ------
\title{Toward Depth Estimation Using Mask-Based Lensless Cameras}

% Single address.
\author{
	\IEEEauthorblockN{M. Salman Asif} \IEEEauthorblockA{Department of 
		Electrical and Computer Engineering \\ University of California, Riverside
		\\ Email: sasif@ece.ucr.edu} }


\maketitle

\setlength{\parskip}{0pt} % 1ex plus 0.5ex minus 0.2ex}
\setlength{\parindent}{2ex}

\begin{abstract}
	Recently, coded masks have been used to demonstrate a thin form-factor lensless camera, FlatCam, in which a mask is placed immediately on top of a bare image sensor. In this paper, we present an imaging model and algorithm to jointly estimate depth and intensity information in the scene from a single or multiple FlatCams. We use a light field representation to model the mapping of 3D scene onto the sensor in which light rays from different depths yield different modulation patterns. We present a greedy depth pursuit algorithm to search the 3D volume and estimate the depth and intensity of each pixel within the camera field-of-view. We present simulation results to analyze the performance of our proposed model and algorithm with different FlatCam settings.

\end{abstract}

\section{Introduction}

% Why lensless: Lenses are thick, bulky, and rigid
Lens-based cameras are standard vision sensors in system that records visual information. However, lens-based cameras are bulky, heavy, and rigid---partly because of the size and material of a lens. The shape of a lens-based camera is also fixed in a cube-like form because of the physical constraints on placing the lens at a certain distance from the sensor. A lensless camera can potentially be very thin and lightweight, can operate over a large spectral range, can provide an extremely wide field of view, and can have curved or flexible shape.

Recently, a new lensless imaging system, called FlatCam, was proposed in \cite{asif2017FlatCam}. FlatCam consists of a coded binary mask placed at a small distance from a bare sensor. 
FlatCam can be viewed as an example of a coded aperture system in which the mask was placed extremely close to the sensor~\cite{fenimore1978coded}. The mask pattern was selected in a way that the image formation model takes a linear separable form. Image reconstruction from the sensor measurements requires solving a linear inverse problem. 

One limitation of the imaging model in \cite{asif2017FlatCam} is that it assumes a 2D scene that consists of a single plane at a fixed distance from the camera. 
In this paper, we present a new imaging model for FlatCam in which the scene consists of multiple planes at different (unknown) depths. We use a light field representation in which light rays from different depths yield different modulation patterns. We use the lightfield representation to analyze the sensitivity of FlatCam to the sampling pattern and depth mismatch. We present a greedy algorithm that jointly estimates the depths and intensity of each pixel. We present simulation results to demonstrate the performance of our algorithm under different settings.

\begin{figure}
	\centering
	\includegraphics[width=1\linewidth]{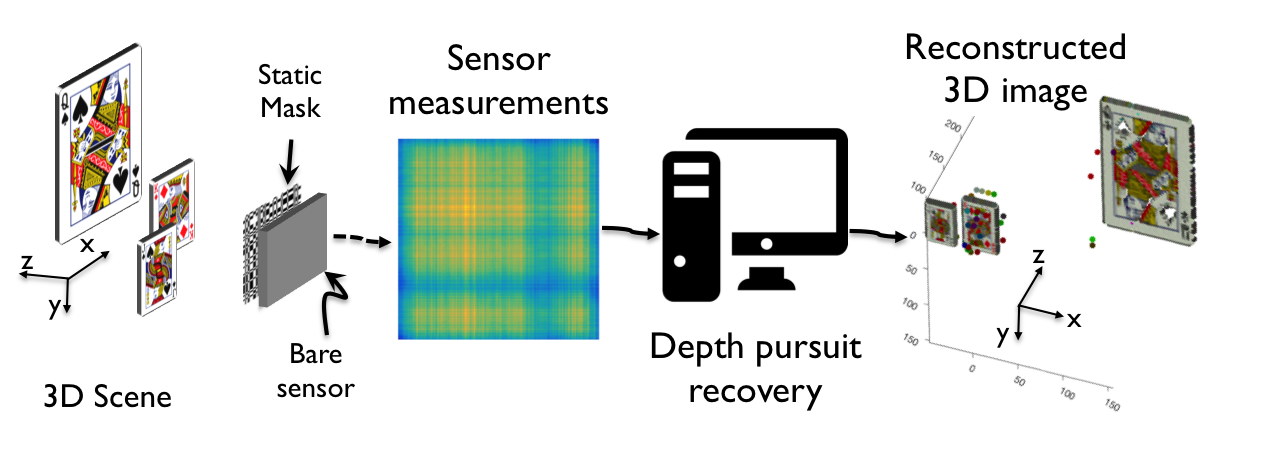}
	\caption{3D imaging with a mask-based lensless camera that consists of a bare sensor with a fixed, binary mask on top of it. Every light source from within the camera field-of-view casts a shadow of the mask on sensor, resulting in a multiplexed image on the sensor. The shadow of any light source depends on its 3D location with respect to the mask-sensor assembly. A depth-selective pursuit algorithm reconstructs the 3D image of the scene.}
	\label{fig:flatcam} 
	\vspace{-10pt}
\end{figure}

\section{Background and related work} 
A pinhole camera is a classical example of a lensless camera in which an opaque mask with a single pinhole is placed in front of a light-sensitive surface. A pinhole camera can potentially take an arbitrary shape. However, a major drawback of a pinhole camera is that it only allows a tiny fraction of the ambient light to pass through the single pinhole; therefore, it typically requires very long exposure times.  
Coded aperture imaging systems extend the idea of a pinhole camera by using a mask with  multiple pinholes \cite{dicke1968scatter,fenimore1978coded,cannon1980coded,durrant1999application,brady2009optical, busboom1998uniformly}. However, the image formed on the sensor is a linear superposition of images from multiple pinholes. We need to solve an inverse problem to recover the underlying scene image from sensor measurements. The primary purpose of the coded aperture is to increase the amount of light recorded at the sensor. 

\begin{figure}
	\centering
	\includegraphics[width=.75\linewidth]{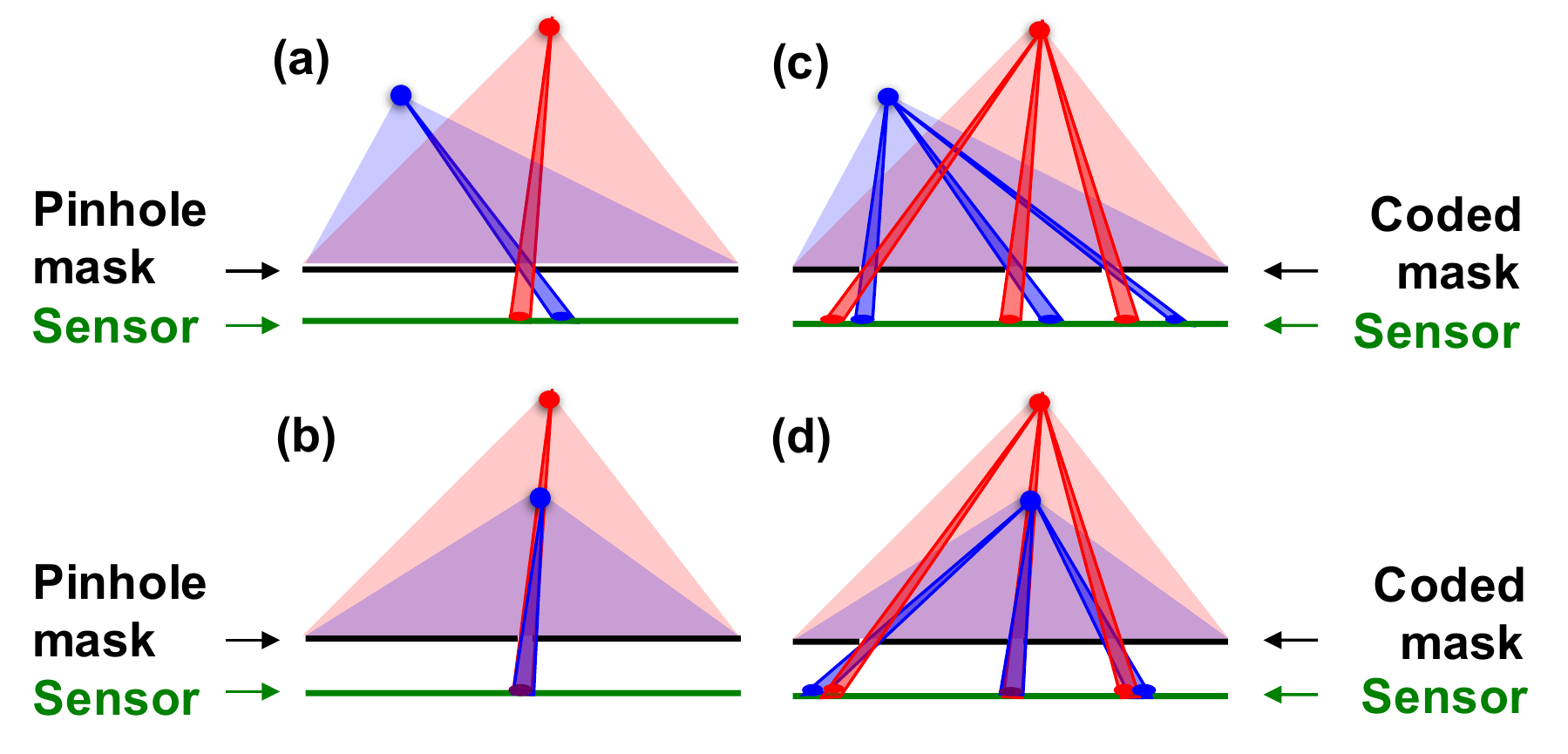}
	\caption{{Examples of imaging with pinhole and coded mask-based cameras. Light rays from all direction hit the mask; rays can only pass through transparent regions (holes). } \textbf{(a,b)} Pinhole cameras preserve angular information but lose depth information as points along the same angle yield identical images, irrespective of their depths. \textbf{(c,d)} Coded aperture-based cameras record coded combination of light from different directions and better preserve depth information.}
	\label{fig:pinhole-to-codedAperture}
	\vspace{-10pt}
\end{figure}

A coded aperture system offers another advantage by virtue of encoding light from different directions and depths differently. Note that a bare sensor can provide the intensity of a light source but not its spatial location. A mask in front of the sensor encodes directional information of the source in  the sensor measurements. 
In a coded aperture system, every light source in the scene casts a unique shadow of the mask onto the sensor. Therefore, sensor measurements encode information about locations and intensities of all the light sources in the scene. 
Consider a single light source with a dark background; the image formed on the sensor will be a shadow of the mask. If we change the angle of the light source, the mask shadow on the sensor will shift. Furthermore, if we change the depth of the light source, the size of the shadow will change (see Figure~\ref{fig:pinhole-to-codedAperture}). 
Thus, we can represent the relationship between all the points in the scene and the sensor measurements as a linear system, which depends on the pattern and the placement of the mask. We can solve this system using an appropriate computational algorithm to recover the image of the scene. 

The depth-dependent imaging capability in coded aperture systems is known since the pioneering work in this domain~\cite{barrett1973fresnelzoneImaging,fenimore1978coded}. The following excerpt in \cite{fenimore1978coded} summarizes it well: "One can reconstruct a particular depth in the object by treating the picture as if it was formed by an aperture scaled to the size of the shadow produced by the depth under consideration." However, the classical methods assume that the scene consists of a single plane at known depth. In this paper, we assume that the depth scene consists of multiple depth planes and the true depth map is unknown at the time of reconstruction. 

Coded-aperture cameras have traditionally been used for imaging wavelengths beyond the visible spectrum (e.g., X-ray and gamma-ray imaging), for which lenses or mirrors are expensive or infeasible \cite{dicke1968scatter,fenimore1978coded,cannon1980coded,durrant1999application,brady2009optical, busboom1998uniformly}. 
Mask-based lensless designs have been proposed for flexible field-of-view selection in \cite{zomet2006lensless}, compressive single-pixel imaging using a transmissive LCD panel \cite{huang2013lensless}, and separable coded masks \cite{deweert2015lensless}. 
In recent years, coded masks and light modulators have been added to lens-based cameras in different configurations to build novel imaging devices that can capture image and depth \cite{levin2007image} or 4D light field \cite{veeraraghavan2007dappled,marwah2013compressive} from a single coded image. Light field imaging by moving a lensless cameras has been demonstrated in \cite{zhang2005light} and 3D imaging with a single-shot diffuser-based lensless camera was recently demonstrated in \cite{antipa2017diffusercam}. 

\section{FlatCam: Replacing lenses with coded masks and computations} 
FlatCam is a coded aperture system that consists of a bare, planar sensor and a binary mask \cite{asif2017FlatCam}. 
Coded-aperture cameras have traditionally been used for imaging wavelengths beyond the visible spectrum (e.g., X-ray and gamma-ray imaging), for which lenses or mirrors are expensive or infeasible \cite{dicke1968scatter,fenimore1978coded,brady2009optical}. 
A bare sensor can provide information about the intensity of a light source but not its spatial location. By adding a mask in front of the sensor, we can encode directional information of the source in  the sensor measurements.

The imaging model in \cite{asif2017FlatCam} assumes that the scene consists of a single plane parallel to mask-sensor planes. Let us consider a 1D imaging system, shown in Fig.~\ref{fig:3Dgeometry}, in which a 1D mask is placed at a distance $d$ in front of a 1D sensor array with $M$ pixels, and the scene consists of a \textit{single} plane at distance $D$ from the sensor with $N$ scene pixels. Let us denote a scene pixel as $l(\theta)$, where $\theta$ is uniformly distributed along an angular interval $[\theta_-,\theta_+]$. We can represent the measurement at sensor pixel $s$ as 
\begin{equation}\label{eq:planarMeasuerments}
	I(s) = \sum_{\theta = \theta_-}^{\theta_+} \varphi(s,\theta) l(\theta),
\end{equation}
where $\varphi(\theta,s)$ denotes the modulation coefficient of the mask for a light ray between scene pixel $l(\theta)$ and the sensor pixel at location $s$. We can write \eqref{eq:planarMeasuerments} in a compact form as 
\begin{equation}
I = \Phi l,
\end{equation}
where $\Phi$ denotes an $M\times N$ system matrix that maps $N$ scene pixels to $M$ sensor pixels.  
This is a linear system that we can solve using an appropriate computational algorithm to recover the image of the scene (more details can be found in \cite{asif2017FlatCam}).

\section{Depth estimation using FlatCams}

\subsection{Imaging model} 
\begin{figure}
	\centering
	\begin{subfigure}[t]{0.475\linewidth}		\includegraphics[width=1\linewidth]{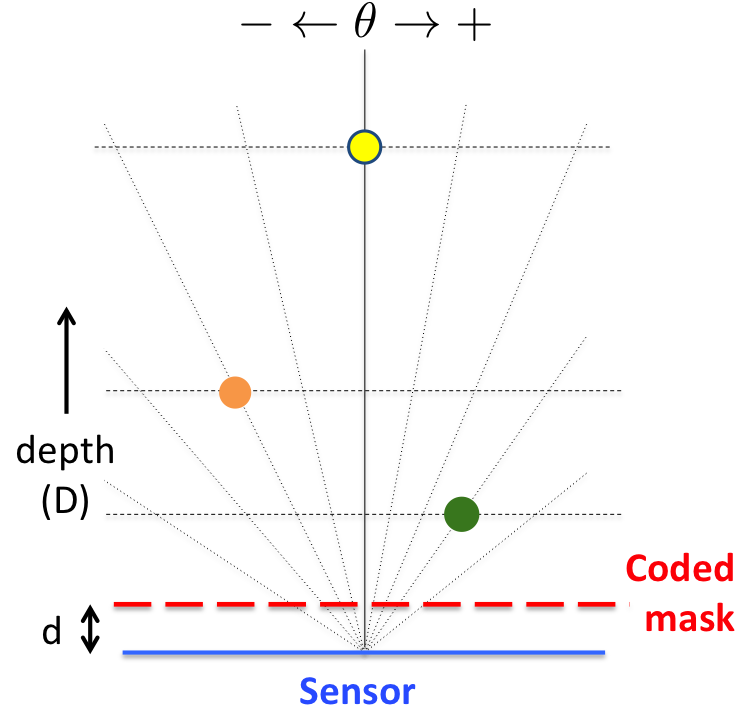}	 
		\caption{\textbf{Imaging system geometry.} Mask and sensor planes are separated by distance $d$. Each point in the scene (at any angle $\theta$ and distance $D$) contributes to the sensor measurements. Our goal is to jointly estimate depth and intensity of each pixel within the field-of-view.} \label{fig:3Dgeometry}
	\end{subfigure}
	\hspace{5pt}
	\begin{subfigure}[t]{0.475\linewidth}
		\includegraphics[width=1\linewidth]{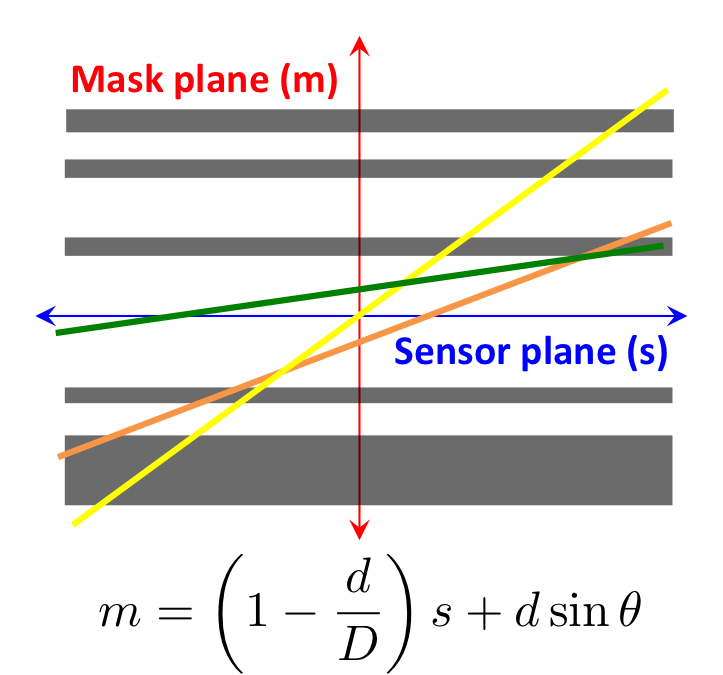}	 
		\caption{\textbf{Lightfield representation of the system.} Angle and depth of a scene point encode intercept and depth of the respective line in lightfield. Horizontal lines denote mask patten. Lightfield is first modulated by the mask pattern and then integrated at the sensor plane. } \label{fig:lightfield}
	\end{subfigure}
	\caption{Geometry of the imaging system in 1D and the corresponding light field representation.}	
	\vspace{-10pt}
\end{figure}

The model in \eqref{eq:planarMeasuerments} assumes a 2D scene that consists of a single plane at some fixed depth. The system matrix $\Phi$ encodes mapping of scene points from that plane to the sensor pixels. In our new model, we consider a 3D scene that consists of multiple planes, each of which contribute to the sensor measurements. Without loss of generality, we consider a 1D imaging system in Fig.~\ref{fig:3Dgeometry} and assume that the sensor plane is centered at the origin and the mask plane is placed in front of it at distance $d$. The scene consists of $K$ planes at depths $[D_1,\ldots,D_K]$ and the scene pixels are distributed uniformly along angles in interval $[\theta_-,\theta_+]$, as before. We can describe the measurement at sensor pixel $s$ as
\begin{equation}\label{eq:depthMeasurements}
I(s) = \sum_{D = D_1}^{D_K} \sum_{\theta = \theta_-}^{\theta_+} \varphi(s,\theta,D) l(\theta,D),
\end{equation}
where $\varphi(s,\theta,D)$ denotes modulation coefficient of the mask for a light ray between light source $l(\theta,D)$ and the sensor pixel at location $s$. We can write 
\eqref{eq:depthMeasurements} in a compact form as 
\begin{equation}
I = \sum_{D=D_1}^{D_K} \Phi_D l_D,
\end{equation}
where each $l_D$ denotes intensity of $N$ pixels in a plane at depth $D$ and $\Phi_D$ is an $M\times N$ matrix that represents the mapping of $l_D$ onto the sensor. 

For the case of 3D imaging, let us represent the light distribution as $L(\theta_x,\theta_y,z)$ for $z>d$. Measurements for a sensor pixel located at $(s_u,s_v)$ can then be described using the following system of linear equations: 
\begin{equation}\small
\begin{aligned}
&I(s_u,s_v) = {} \sum_{\theta_x,\theta_y,z} L(\theta_x,\theta_y,z) \times \\ &\text{mask}\left[\left(1-\frac{d}{z}\right)s_u+d\sin\theta_x,\left(1-\frac{d}{z}\right)s_v+d\sin\theta_y\right],\label{eq:3Dsampling}
\end{aligned}
\end{equation}
where $\text{mask}[u,v]$ denotes the transparency value of the mask at location $(u,v)$ within the mask plane. 
If the mask pattern is symmetric and separable in $(\theta_x, \theta_y)$ space, then we can write the 2D measurements in \eqref{eq:3Dsampling} as 
\begin{equation}\label{eq:3Dseparable}
I = \sum_{D=D_1}^{D_K} \Phi_D L_D \Phi_D^T, 
\end{equation}
where $\Phi_D$ is an $M\times N$ matrix and $L_D$ denotes $N\times N$ light distribution corresponding to plane at depth $D$. 

Furthermore, we can include multiple cameras at different locations and orientations with respect to a reference frame. Such a system will provide us multiple sensor measurements of the form 
\begin{equation}\label{eq:depthMultipleCam}
I_c = \sum_{D=D_1}^{D_K} \Phi_{D,c} L_D  \Phi_{D,c}^T,
\end{equation}
where $I_c$ denotes sensor measurements at camera $c$ and $\Phi_{D,c}$ is matrix that represents mapping of $L_D$ onto camera $c$. 

\subsection{Joint image and depth reconstruction} 
We estimate the depth and intensity of each pixel within the field-of-view of our cameras using a \textit{greedy depth-selective algorithm}, in which we assume a sparse prior that $L(\theta_x,\theta_y,D)$ has nonzero value only for one depth. 
Our proposed algorithm is inspired by structured sparse recovery algorithms in model-based compressive sensing~\cite{baraniuk2010model}. 
% We also consider occlusion of the points and give a higher preference for planes closer to the sensor. 

To simplify the presentation, let us represent \eqref{eq:3Dseparable} or  \eqref{eq:depthMultipleCam} as the following general linear system: 
\begin{equation}
\mathcal{I} = \mathcal{A}(L),
\end{equation}
where $L$ is an $N\times N \times K$ light distribution with $N\times N$ spatial resolution and $K$ depth planes, $\mathcal{A}$ denotes the linear measurement operator in \eqref{eq:3Dseparable} or \eqref{eq:depthMultipleCam}, and $\mathcal{I}$ denotes the sensor measurements. 

Suppose our current estimate of $L$ is $\tilde L$ with exactly one depth assigned to each pixel. Let us denote the initial depth map as $\Omega$. In our experiments, we initialize the depth estimate with the farthest plane in the scene. Our proposed \textit{depth-selective pursuit} algorithm is an iterative method that performs the following three main steps at every iteration: \\
{\bf Compute proxy depth estimate.} We first select new candidate depth for each pixel by picking the maximum magnitude in the following proxy map corresponding to each pixel: $\mathcal{P} = \mathcal{A}^T[\mathcal{I}-\mathcal{A}(\tilde L)]$. Let us denote the new depth map as $\tilde \Omega$. \\
{\bf Merge depths and estimate image. } We first merge the original depth estimate with the proxy depth estimate. Let us denote the merged depth support as $T = \{\Omega \cup \tilde \Omega \}$. Then we solve a least-squares problem over the merged depth support as $\hat L = \arg\,\min_{L} \|\mathcal{I}_T - \mathcal{A}_T L\|_2^2$. \\
{\bf Prune depth and threshold image.} We prune the depth estimate at every spatial location by picking the depth corresponding to higher pixel intensity in $\hat L$. Finally, we threshold $\hat L$ to $\tilde L$ that has only one nonzero depth per spatial location. 
%

% A simulation result of our algorithm is given in Fig.~\ref{fig:results}. 

% \if 1 

\subsection{Depth sampling and sensitivity}
For a single camera, sensor measurements for a single point source at location $(\theta,D)$ can be described as 
\begin{equation}
I(s) \propto \text{mask}\left[\left(1-\frac{d}{D}\right) s + d\sin \theta\right].
\end{equation}
From this lightfield expression we note that the slope of a line corresponding to any light source is inversely proportional to its depth. As a light source moves farther from the mask-sensor assembly, its line would rotate around the center. As a light source moves along a plane at a fixed depth, its line would shift with the same slope. Therefore, in our imaging model, we select depth planes in a given range by sampling lightfield at uniform angles, which results in planes at non-uniform depths. 

% We use this relation to divide the scene into planes at non-uniform depths and uniform angles to model the 3D space within camera field-of-view. 
% The points at the same depth and different angles cast shadows that are shifted versions of one another. Points along the same angle with respect to the center of the sensor cause the same shift in the mask shadow. 

% \fi

\section{Experimental results}
To validate the performance of our proposed imaging model and reconstruction algorithm, we performed extensive simulations under different settings of FlatCam parameters.

\begin{figure}[!h]
	\centering
	\begin{subfigure}[t]{0.23\textwidth}
		\includegraphics[width=\linewidth]{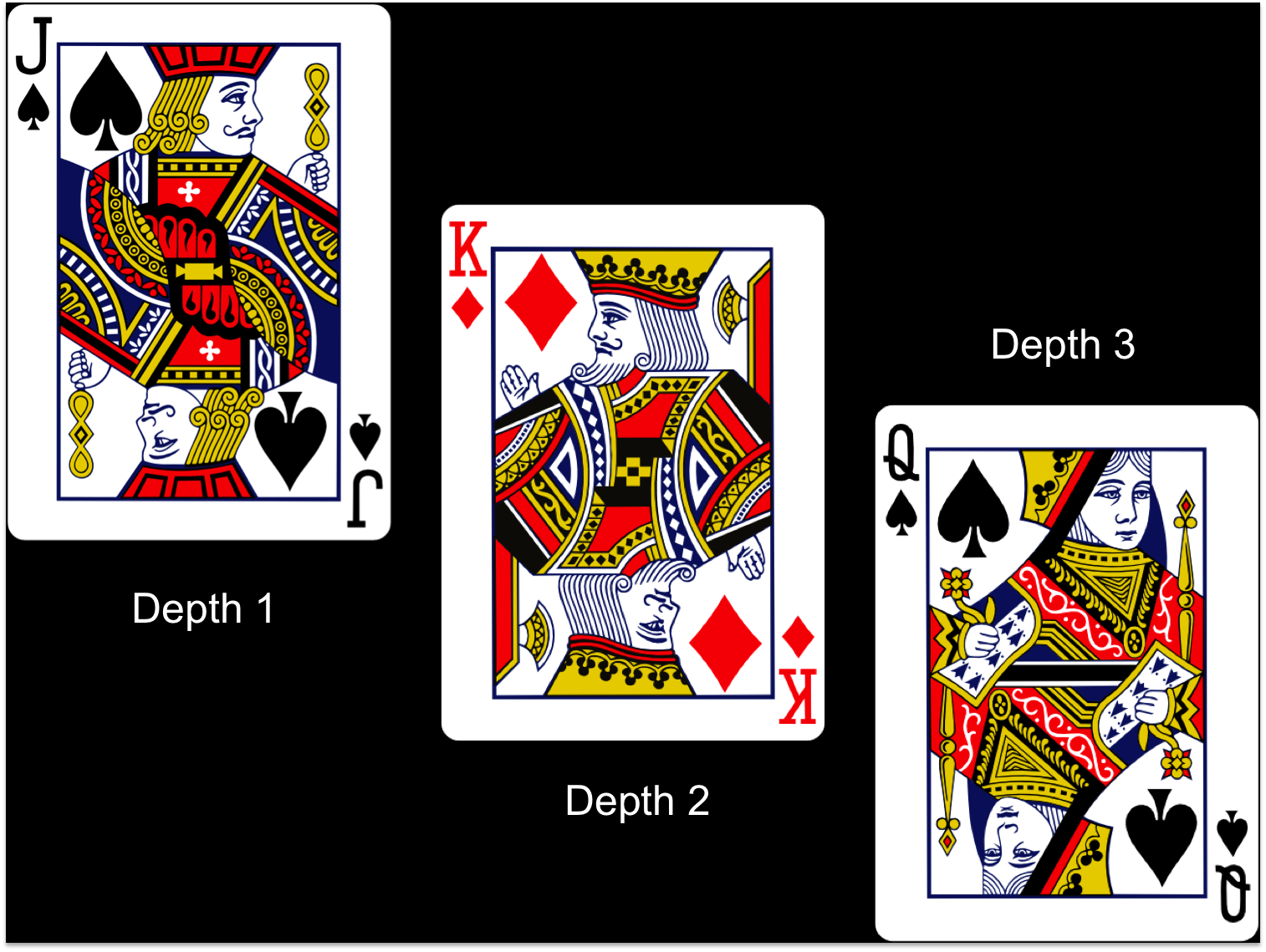}
		\caption{Test scene with three cards placed at three different depths picked out of $K=10$ depth planes at random.}
		\label{fig:testImage}
	\end{subfigure}
	\hspace{2pt}
	\begin{subfigure}[t]{0.23\textwidth}
		\includegraphics[width=\linewidth]{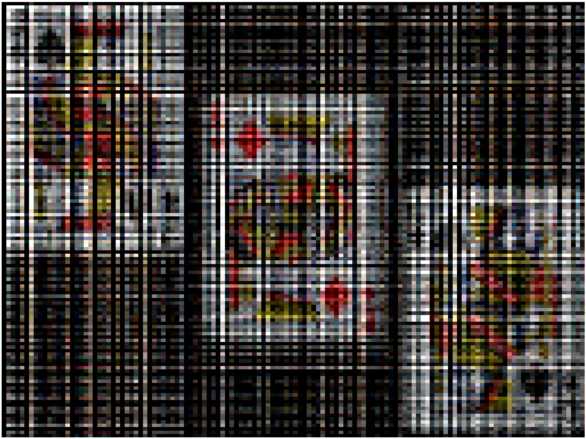}
		\caption{Image reconstructed by assuming that the scene consists of a single plane at a fixed depth. \\ PSNR = 15\,dB}
		\label{fig:testImage}
	\end{subfigure}
	\\ \vspace{10pt}
	\begin{subfigure}[t]{0.22\textwidth}
		\includegraphics[width=\linewidth]{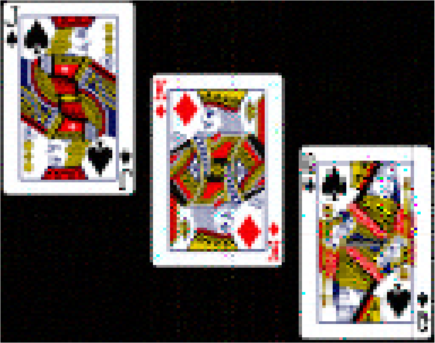}
		\caption{Image reconstructed using depth-selective algorithm that jointly estimates the depth and intensity of each pixel. \\PSNR = 33.5\,dB}
		\label{fig:testImage}
	\end{subfigure}
	\hspace{2pt}
	\begin{subfigure}[t]{0.22\textwidth}
		\includegraphics[width=\linewidth]{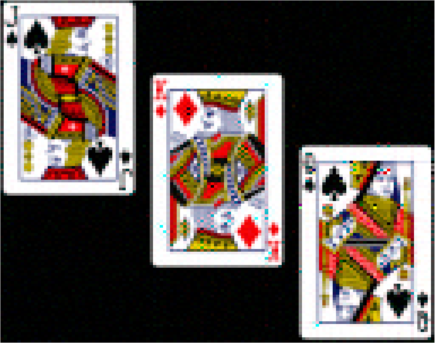}
		\caption{Image reconstructed with three cameras via depth-selective algorithm that jointly estimates the depth and intensity of each pixel. \\ PSNR = 39\,dB}
		\label{fig:testImage}
	\end{subfigure}
	\vspace{5pt}
	\caption{Simulation results to demonstrate the effect of depth sensitivity and a result for our joint depth and intensity reconstruction algorithm. }	\label{fig:results}
	\vspace{-10pt}
\end{figure}

First, we show results of a simple simulation in which our scene consists of three depth planes as shown in Fig.~\ref{fig:results}. We simulated an imaging system in which a binary mask is placed at 1mm distance from the sensor. We simulated a 3D voxel space with ten depth planes within a depth range of 100mm and 3m. We chose the ten depth planes by uniformly sampling the lightfield representation. We simulated a scene with $128\times 128$ spatial resolution, a mask with a binary random sequence, and a sensor with $256\times 256$ pixels. To generate sensor measurements, we assumed that the 3D scene consists of three planes chosen at random out of ten fixed planes. We added a small amount of Gaussian noise in the true measurements. To reconstruct the 3D scene, we solved the depth-selective pursuit algorithm described in the previous section. The results are presented in Fig.~\ref{fig:results}, where (a) denotes pixel intensities for three planes in a 3D scene, (b) denotes image reconstructed by assuming that all pixels belong to a plane at fixed depth, and (c) denotes images reconstructed by solving the depth-pursuit algorithm on the same measurements.

Next we discuss an experiment that demonstrates the robustness of our proposed model and method against mismatch in the locations of the original depth planes and those used for reconstruction. In this experiment, we simulated imaging system with $K = \{5, 10, 15, 20, 25\}$ depth planes chosen uniformly in the lightfield representation. We selected the three depth planes for the scene at random and calculated the peak signal to noise ratio (PSNR) for the recovered images. We present PSNR for each test (averaged over ten instances) in Table~\ref{table:psnr}. We see that the quality of reconstruction remains almost the same as we increase the number of depth planes in our model. The computational complexity, however, slightly increases as we increase the number of depth planes. 

Finally, we present an experiment in which we simulated a system with three lensless cameras in a convex  geometry, where one camera is used as a reference to generate depth planes in the scene. The other two cameras see tilted planes in their field of view. An advantage of such a configuration is that the depth information of the pixels is converted into angular information. However, this configuration also makes a strong assumption that the scene only consists of the finite number of tilted planes. The simulation results of three camera system are also summarized in Table~\ref{table:psnr}. An example of image reconstruction for this case is shown in Fig.~\ref{fig:results}(d). 

% - table with psnr performance for different images under different number of depth planes and scene complexity 
% - oversampling in spatial vs depth calibration

\renewcommand{\arraystretch}{1.5}

\begin{table}
	\centering
	\begin{tabular}{c|c|c|c|c|c|}
		\cline{2-6}
		& \multicolumn{5}{ c| }{\textbf{Number of imaging depth planes ($K$)}} \\ \cline{2-6}
		& 5 & 10 & 15 & 20 & 25\\ \cline{1-6}
		\multicolumn{1}{ |c|  }{\bf Single camera} & 33.83 & 33.58 & 31.27 & 30.5 & 30.99 \\ \cline{1-6}
		\multicolumn{1}{ |c|  }{\bf Three cameras} & 39.07 & 39.09 & 40 & 39.54 & 39.58 \\ \cline{1-6}
	\end{tabular}
	\caption{PSNR (in dB) comparison of reconstructed images for different number of depth planes and different number of cameras.}\label{table:psnr}
\end{table}

 \balance
{\footnotesize
	\bibliographystyle{IEEEtran}
	\bibliography{flatcam}

\begin{thebibliography}{10}
\providecommand{\url}[1]{#1}
\csname url@samestyle\endcsname
\providecommand{\newblock}{\relax}
\providecommand{\bibinfo}[2]{#2}
\providecommand{\BIBentrySTDinterwordspacing}{\spaceskip=0pt\relax}
\providecommand{\BIBentryALTinterwordstretchfactor}{4}
\providecommand{\BIBentryALTinterwordspacing}{\spaceskip=\fontdimen2\font plus
\BIBentryALTinterwordstretchfactor\fontdimen3\font minus
  \fontdimen4\font\relax}
\providecommand{\BIBforeignlanguage}[2]{{%
\expandafter\ifx\csname l@#1\endcsname\relax
\typeout{** WARNING: IEEEtran.bst: No hyphenation pattern has been}%
\typeout{** loaded for the language `#1'. Using the pattern for}%
\typeout{** the default language instead.}%
\else
\language=\csname l@#1\endcsname
\fi
#2}}
\providecommand{\BIBdecl}{\relax}
\BIBdecl

\bibitem{asif2017FlatCam}
M.~S. Asif, A.~Ayremlou, A.~Sankaranarayanan, A.~Veerarghavan, and R.~Baraniuk,
  ``Flatcam: Thin, bare-sensor cameras using coded aperture and computation,''
  \emph{IEEE Transactions on Computational Imaging}, vol.~3, no.~3, pp.
  384--397, Sept 2017.

\bibitem{fenimore1978coded}
E.~Fenimore and T.~Cannon, ``Coded aperture imaging with uniformly redundant
  arrays,'' \emph{Applied optics}, vol.~17, no.~3, pp. 337--347, 1978.

\bibitem{dicke1968scatter}
R.~Dicke, ``Scatter-hole cameras for x-rays and gamma rays,'' \emph{The
  Astrophysical Journal}, vol. 153, p. L101, 1968.

\bibitem{cannon1980coded}
T.~Cannon and E.~Fenimore, ``Coded aperture imaging: Many holes make light
  work,'' \emph{Optical Engineering}, vol.~19, no.~3, pp. 193--283, 1980.

\bibitem{durrant1999application}
P.~Durrant, M.~Dallimore, I.~Jupp, and D.~Ramsden, ``The application of pinhole
  and coded aperture imaging in the nuclear environment,'' \emph{Nuclear
  Instruments and Methods in Physics Research Section A: Accelerators,
  Spectrometers, Detectors and Associated Equipment}, vol. 422, no.~1, pp.
  667--671, 1999.

\bibitem{brady2009optical}
D.~J. Brady, \emph{Optical imaging and spectroscopy}.\hskip 1em plus 0.5em
  minus 0.4em\relax John Wiley \& Sons, 2009.

\bibitem{busboom1998uniformly}
A.~Busboom, H.~Elders-Boll, and H.~Schotten, ``Uniformly redundant arrays,''
  \emph{Experimental Astronomy}, vol.~8, no.~2, pp. 97--123, 1998.

\bibitem{barrett1973fresnelzoneImaging}
\BIBentryALTinterwordspacing
H.~H. Barrett and F.~A. Horrigan, ``Fresnel zone plate imaging of gamma rays;
  theory,'' \emph{Appl. Opt.}, vol.~12, no.~11, pp. 2686--2702, Nov 1973.
  [Online]. Available: \url{http://ao.osa.org/abstract.cfm?URI=ao-12-11-2686}
\BIBentrySTDinterwordspacing

\bibitem{zomet2006lensless}
A.~Zomet and S.~K. Nayar, ``Lensless imaging with a controllable aperture,'' in
  \emph{IEEE Computer Society Conference on Computer Vision and Pattern
  Recognition}, vol.~1, 2006, pp. 339--346.

\bibitem{huang2013lensless}
G.~Huang, H.~Jiang, K.~Matthews, and P.~Wilford, ``Lensless imaging by
  compressive sensing,'' in \emph{20th IEEE International Conference on Image
  Processing}, 2013, pp. 2101--2105.

\bibitem{deweert2015lensless}
M.~J. DeWeert and B.~P. Farm, ``{Lensless coded-aperture imaging with separable
  doubly-{Toeplitz} masks},'' \emph{Optical Engineering}, vol.~54, no.~2, pp.
  023\,102--023\,102, 2015.

\bibitem{levin2007image}
A.~Levin, R.~Fergus, F.~Durand, and W.~T. Freeman, ``Image and depth from a
  conventional camera with a coded aperture,'' in \emph{ACM Transactions on
  Graphics (TOG)}, vol.~26, no.~3.\hskip 1em plus 0.5em minus 0.4em\relax ACM,
  2007, p.~70.

\bibitem{veeraraghavan2007dappled}
A.~Veeraraghavan, R.~Raskar, A.~Agrawal, A.~Mohan, and J.~Tumblin, ``Dappled
  photography: Mask enhanced cameras for heterodyned light fields and coded
  aperture refocusing,'' \emph{ACM Transactions on Graphics (TOG)}, vol.~26,
  no.~3, p.~69, 2007.

\bibitem{marwah2013compressive}
K.~Marwah, G.~Wetzstein, Y.~Bando, and R.~Raskar, ``Compressive light field
  photography using overcomplete dictionaries and optimized projections,''
  \emph{ACM Transactions on Graphics (TOG)}, vol.~32, no.~4, p.~46, 2013.

\bibitem{zhang2005light}
C.~Zhang and T.~Chen, ``Light field capturing with lensless cameras,'' in
  \emph{IEEE International Conference on Image Processing (ICIP)},
  vol.~3.\hskip 1em plus 0.5em minus 0.4em\relax IEEE, 2005, pp. III--792.

\bibitem{antipa2017diffusercam}
N.~Antipa, G.~Kuo, R.~Heckel, B.~Mildenhall, E.~Bostan, R.~Ng, and L.~Waller,
  ``Diffusercam: Lensless single-exposure 3d imaging,'' \emph{arXiv preprint
  arXiv:1710.02134}, 2017.

\bibitem{baraniuk2010model}
R.~G. Baraniuk, V.~Cevher, M.~F. Duarte, and C.~Hegde, ``Model-based
  compressive sensing,'' \emph{IEEE Transactions on Information Theory},
  vol.~56, no.~4, pp. 1982--2001, 2010.

\end{thebibliography}
}

\end{document}